\documentclass[11pt]{article}
\usepackage{acl2016}
\usepackage{latexsym}
\usepackage{url}
\usepackage{times}
\usepackage{graphicx}
\usepackage{algorithm}
\usepackage{extarrows}
\usepackage{multirow}
\usepackage{multicol}
\usepackage{bm}
\usepackage{algpseudocode}
\usepackage{indentfirst}
\usepackage{amsfonts,amssymb}
\usepackage{amsmath}
\usepackage{amsthm}
\usepackage{array}
\usepackage{subcaption}
\usepackage{float}

\title{Neural Headline Generation with Sentence-wise Optimization}

\author{Ayana, Shiqi Shen, Yu Zhao, Zhiyuan Liu\thanks{Corresponding author: Z. Liu (liuzy@tsinghua.edu.cn)}, Maosong Sun\\
	 Department of Computer Science and Technology, \\
	State Key Lab on Intelligent Technology and Systems, \\
	National Lab for Information Science and Technology, Tsinghua University, Beijing, China \\
}

\date{}

\begin{document}
	\maketitle
	
\begin{abstract}
	
Recently, neural models have been proposed for headline generation by learning to map documents to headlines with recurrent neural networks. Nevertheless, as traditional neural network utilizes maximum likelihood estimation for parameter optimization, it essentially constrains the expected training objective within word level rather than sentence level. Moreover, the performance of model prediction significantly relies on training data distribution. To overcome these drawbacks, we employ minimum risk training strategy in this paper, which directly optimizes model parameters in sentence level with respect to evaluation metrics and leads to significant improvements for headline generation. Experiment results show that our models outperforms state-of-the-art systems on both English and Chinese headline generation tasks.
	
\end{abstract}
	
\section{Introduction}

Automatic text summarization is the process of creating a coherent, informative and brief summary for a document. 
Text summarization is expected to understand the main theme of the documents and then output a condensed summary contains as many key points of the original document as it can within  a length limit. Text summarization approaches can be divided into two typical categories: extractive and generative. Most extractive summarization systems simply select a subset of existing sentences from original documents as summary. Despite of its simplicity, extractive summarization has some intrinsic drawbacks, e.g., unable to generate coherent and compact summary in arbitrary length or shorter than one sentence.

In contrast, generative summarization builds semantic representation of a document and creates a summary with sentences not necessarily presenting in the original document  explicitly. When the generated summary is required to be a single compact sentence, we name the summarization task as \emph{headline generation} \cite{dorr2003hedge}.
Most previous works heavily rely on modeling latent linguistic structures of input document, via syntactic parsing and semantic parsing, which always bring inevitable errors and degrade summarization quality.

Recent years have witnessed great success of deep neural models for various natural language processing tasks \cite{cholearning,sutskever2014sequence,bahdanau2015neural,ranzato2015sequence} including text summarization. Taking neural headline generation (NHG) for example, it learns to build a large neural network, which takes a document as input and directly outputs a compact sentence as headline of the document. Compared with conventional generative methods, NHG exhibits the following advantages: (1) NHG is fully data-driven, requiring no linguistic information. (2) NHG is completely end-to-end, which does not explicitly model latent linguistic structures, and thus prevents error propagation. Moreover, the attention mechanism \cite{bahdanau2015neural} is introduced in NHG, which learns a soft alignment over input document to generate more accurate headline \cite{rush2015neural}.

Nevertheless, NHG still confronts a significant problem: current models are mostly optimized at the word level instead of sentence level, which prevents them from capturing various aspects of summarization quality.
In fact, it is essentially desirable to incorporate the implicit sentence-wise information contained in the evaluation criteria, e.g. ROUGE, into NHG model.

To address this issue, we propose to apply the \emph{minimum Bayes risk} technique in tuning NHG model with respect to evaluation metrics. Specifically, we utilize \emph{minimum risk training} (MRT), which aims at minimizing a sentence-wise loss fuction over the training data. To the best of our knowledge, although MRT has been widely used in many NLP tasks such as statistical machine translation \cite{och2003minimum,smith2006minimum,gao2014learning,shen2015minimum}, it has not been well considered in the research of text summarization.

We conduct experiments on three real-world datasets in English and Chinese respectively. Experiment results show that, NHG with MRT can significantly and consistently improve the summarization performance as compared to NHG with MLE, and other baseline systems. Moreover, we explore the influence of employing different evaluation metrics and find the superiority of our model stable in MRT.
	
\section{Background}
\label{sec:model}
\begin{figure}[!t]
	\centering
	\includegraphics[width=6.5cm,angle=-90]{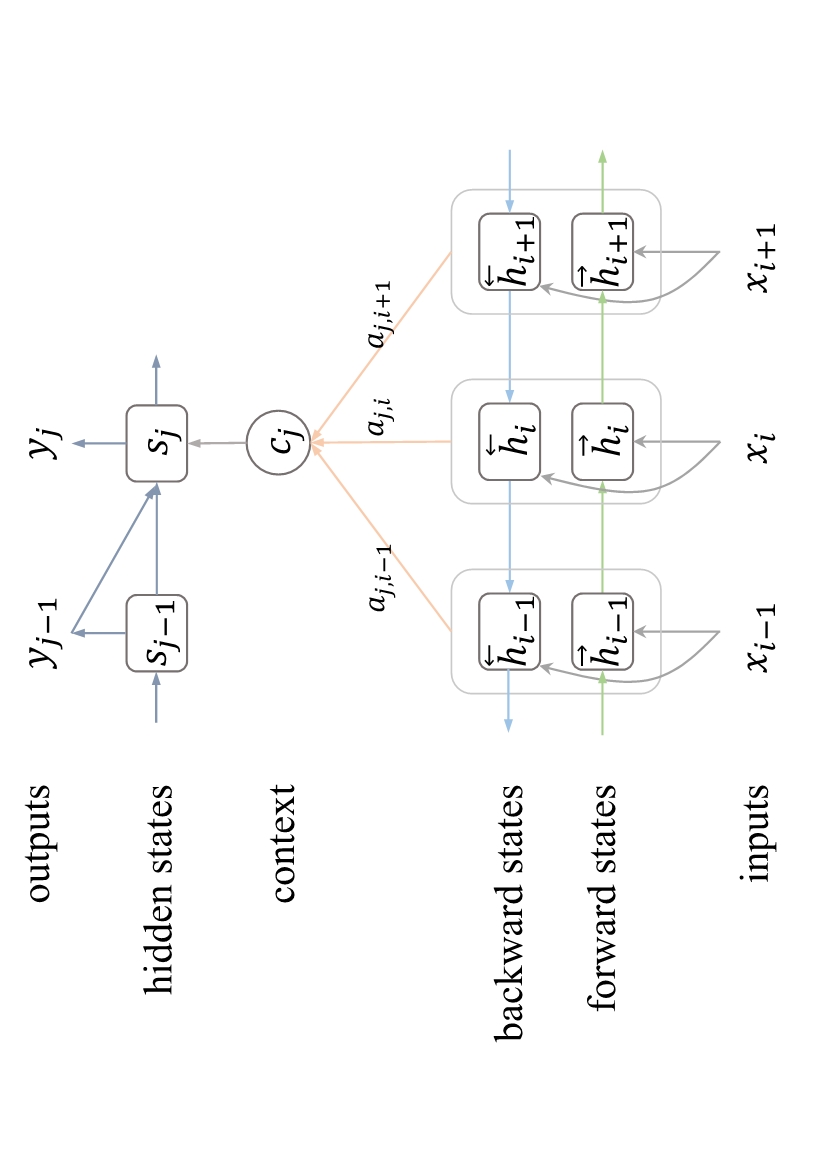}
	\caption{The framework of NHG.}
	\label{fig:net}
\end{figure}

In this section, we formally define the problem of neural headline generation and introduce the notations used in our model. Denote the input document $\mathbf{x}$ as a sequence of words $\{\mathbf{x}_{1}, \cdots, \mathbf{x}_{M}\}$, where each word $\mathbf{x}_{i}$ comes from a fixed vocabulary $V$. Headline generator aims to take $\mathbf{x}$ as input, and generates a short headline $\mathbf{y} = \{\mathbf{y}_{1}, \cdots, \mathbf{y}_{N}\}$ with length $N < M$, such that the conditional probability of $\mathbf{y}$ given $\mathbf{x}$ is maximized. The log conditional probability can be further formalized as:
\begin{equation}
	\log \Pr(\mathbf{y}|\mathbf{x}; \theta) = \sum_{j=1}^{N} \log \Pr(\mathbf{y}_{j}|\mathbf{x},\mathbf{y}_{<j};\theta),
	\label{eq:define}
\end{equation}
where $\mathbf{y}_{<j} = \{\mathbf{y}_1, \ldots, \mathbf{y}_{j-1}\}$, $\theta$ indicates model parameters. That is, the $j$-th word $\mathbf{y}_{j}$ in headline is generated according to all $\mathbf{y}_{<j}$ generated in past and the input document $\mathbf{x}$. In NHG, we adopt an encoder-decoder framework to parameterize $\Pr(\mathbf{y}_{j}|\mathbf{x},\mathbf{y}_{<j};\theta)$, as shown in Fig. \ref{fig:net}.

The encoder of the model encodes the input document $\mathbf{x}$ into low-dimensional vectors $(\mathbf{h}_{1}, \dots, \mathbf{h}_{i}, \dots, \mathbf{h}_{M})$ using bi-directional recurrent neural network with GRU units, where $\mathbf{h}_{i}$ is the concatenation of forward and backward states corresponding to word $x_{i}$. Then, the decoder sequentially generates headline words based on these vectors and decoder hidden states, using a uni-directional GRU recurrent neural network with attention, i.e.,
\begin{equation}
	\Pr(\mathbf{y}_{j}|\mathbf{x},\mathbf{y}_{<j};\theta) = \Pr(\mathbf{y}_{j}|\mathbf{c}_ j,\mathbf{s}_j, \mathbf{y}_{j-1}, \theta),
	\label{eq:generate}
\end{equation}
where $\mathbf{c}_ j$ stands for the context for generating the $j$-th headline word and is calculated utilizing the attention mechanism. $\mathbf{s}_j$ is the $j$-th hidden state of the decoder, and $\theta$ denotes a set of model parameters. Please refer to~\cite{bahdanau2015neural,sutskever2014sequence} for more details. 

\section{Minimum Risk Training for NHG}

Given a dataset $D$ with large-scale document-headline pairs $\{(\mathbf{x}^{(1)}, \mathbf{y}^{(1)}), \ldots, (\mathbf{x}^{(T)}, \mathbf{y}^{(T)})\}$, we propose to use minimum risk training to optimize model parameters instead of the conventional maximum likelihood estimation. We employ the famous ROUGE evaluation metrics to compose the expected loss function. In this section, we introduce our basic idea in detail.

\subsection{Minimum Risk Training}

In the traditional training strategy, the optimized NHG model parameters are estimated by maximizing the log likelihood of generated headlines over training set $D$:

\begin{equation}
	\mathcal{L}_{\text{MLE}}(\theta) = \sum_{(\mathbf{x}, \mathbf{y}) \in D} \log \Pr(\mathbf{y}|\mathbf{x};\theta).
\end{equation}
	
According to Eq. (\ref{eq:define}), the training procedure is fundamentally maximizing the probability of each word in headline step by step, which will inevitably lose global information. Moreover, $\mathbf{y}_{<j}$ are authentic words from reference headline in the training phase, while in the training phase they are predicted by model. It will lead to error propagation and inaccurate headline generation.

In order to tackle these problems, we propose to use minimum risk training (MRT) strategy. Given a document $\mathbf{x}$, we define $\mathcal{Y} (\mathbf{x};\theta)$ as the set of all possible headlines generated with parameters $\theta$. Regarding $\mathbf{y}'$ as the reference headline of $\mathbf{x}$, we denote $\Delta(\mathbf{y'}, \mathbf{y})$ as the distance between $\mathbf{y}$ and a generated headline $\mathbf{y'}$. MRT defines the objective loss function as follows:

\begin{equation}
	\mathcal{L}_{\text{MRT}}(\theta) = \sum_{(\mathbf{x}, \mathbf{y}) \in D} \mathbb{E}_{\mathcal{Y} (\mathbf{x};\theta)} \Delta(\mathbf{y'}, \mathbf{y}).
\end{equation}
Here $\mathbb{E}_{\mathcal{Y} (\mathbf{x};\theta)}$ indicates the expectation over all possible headlines. Thus the objective function of MRT can be further formalized as:
\begin{equation}
	\mathcal{L}_{\text{MRT}}(\theta) = \sum_{(\mathbf{x}, \mathbf{y}) \in D} \sum_{\mathbf{y'} \in \mathcal{Y}(\mathbf{x};\theta)} \Pr(\mathbf{y'}|\mathbf{x};\theta) \Delta(\mathbf{y'}, \mathbf{y}).
\end{equation}

In this way, MRT manages to minimize the expected loss by perceiving the distance as a measure of assessing the overall risk. Nevertheless, it is usually time-consuming and inefficient to enumerate all possible instances. For simplicity, we draw a subset of samples $\mathcal{S}(\mathbf{x};\theta) \subset \mathcal{Y}(\mathbf{x};\theta)$ from the current probability distribution of generated headlines. The loss function can be approximated as:

\begin{multline}\small
\mathcal{L}_{\text{MRT}}(\theta) =  \\ \sum_{(\mathbf{x}, \mathbf{y}) \in D} \sum_{\mathbf{y'} \in \mathcal{S}(\mathbf{x};\theta)} \frac{\Pr(\mathbf{y'}|\mathbf{x};\theta)^{\epsilon}}{\sum_{\mathbf{y^{*}} \in \mathcal{S}(\mathbf{x};\theta)}^{}\Pr(\mathbf{y^{*}}|\mathbf{x};\theta)^{\epsilon}} \Delta(\mathbf{y'}, \mathbf{y}),
\label{eq:loss}
\end{multline}
where $\epsilon$ is a hyper-parameter that controls the smoothness of the objective function \cite{och2003minimum}. A proper $\epsilon$ value can significantly enhance the effectiveness of MRT. In the experiment, we set $\epsilon$ to $5\times 10^{-3}$. 

\subsection{ROUGE}
\label{sec:rouge}
MRT exploits the distance between two sentences to compose the loss function, which enables us to directly optimize NHG model with respect to a specific evaluation metric of the task. As we know, the most widely used evaluation metric for document summarization is ROUGE \cite{lin2004rouge}. The basic idea of ROUGE is to count the number of overlapping units between computer-generated summaries and the reference summaries, such as overlapped n-grams, word sequences, and word pairs. It is the most common evaluation metric in Document Understanding Conference (DUC), a large-scale summarization evaluation sponsored by NIST \cite{lin2004rouge}. 
When training English models, we adopt negative recall value of ROUGE-1,2 and L to compose $\Delta(\mathbf{y'},\mathbf{y})$. For Chinise models, we utilize negative F1 value of ROUGE-1,2 and L to compose $\Delta(\mathbf{y'},\mathbf{y})$. 
	
\section{Experiments}

In this paper, we evaluate our methodology on both English and Chinese headline generation tasks. We first introduce the datasets and evaluation metrics used in the experiment. Then we demonstrate that our model performs the best compared with state-of-the-art baseline systems. We also analyze the influence of different parameters in detail to gain more insights.

\subsection{Datasets and Evaluation Metrics}
\label{sec:enData}

\textbf{English Datasets}. In the experiment, we utilize the English Gigaword Fifth Edition \cite{parker2011english} corpus, containing $10$ million news articles\footnote{To avoid noises in articles and headlines that may influence the performance, we filter out headlines with bylines, extraneous editing marks and question marks. For English dataset, we preprocess the corpus with tokenization and lower-casing.} with corresponding headlines. We follow the experimental settings in ~\cite{rush2015neural} to collect $4$ million article-headline pairs as the training set. 

We use the dataset from DUC-2004 Task-1 as our test set. It consists of $500$ news articles, each of which is paired with four human-generated reference headlines. \footnote{The dataset can be downloaded from \url{http://duc.nist.gov/} with agreements.} We also take Gigaword test set~\footnote{This dataset is provided by ~\cite{rush2015neural}.} to evaluate our models.
We use the DUC-2003 evaluation dataset of size $624$ as development set to tune the hyper-parameters in MRT.

\textbf{Chinese Dataset}. We conduct experiments on the Chinese LCSTS dataset \cite{lcsts}, consisting of article-headline pairs extracted from Sina Weibo \footnote{The website of Sina Weibo is \url{http://weibo.com/}. A typical news article posted in Weibo is limited to $140$ Chinese characters, and the corresponding headline is usually set in a pair of square brackets at the beginning of the news article.}. LCSTS are composed of three parts, containing $2.4$ million, $10,666$ and $1,106$ article-headline pairs respectively. Those pairs in Part-II and Part-III are labeled with relatedness scores by human annotation that indicate how relevant an article and its headline are \footnote{Each pair in Part-II is labeled by only one annotator, and in Part-III is by three annotators.}. In the experiment, we take Part-I of LCSTS as training set, Part-II as development set and Part-III as test set. We only reserve those pairs with scores no less than $3$. It is worth mentioning that, we take Chinese characters as inputs of NHG instead of words in order to prevent the influence of Chinese word segmentation errors. In addition, we replace all digits with \# for both English and Chinese corpus.

\textbf{Evaluation metrics}. In the experiment, we use ROUGE~\cite{lin2004rouge}, as introduced in Section \ref{sec:rouge}, to evaluate the performance of headline generation. 


Following ~\cite{rush2015neural,rush2016,CoRR}, for DUC2003 and DUC2004, we report recall scores of ROUGE-1($R1_{R}$) , ROUGE-2($R2_{R}$) and ROUGE-L($RL_{R}$) with official 75 bytes ceiling limit. And following \cite{rush2015neural,rush2016,CoRR}, for Gigaword test set, we report full-length F1 scores of ROUGE-1($R1_{F1}$) , ROUGE-2($R2_{F1}$) and ROUGE-L($RL_{F1}$). Since a shorter summary tends to get lower recall score, when testing on DUC datasets, we set the minimum length of a generated headline as 10. Note that we report 75 bytes capped recall scores only. In this case summaries that longer than 75 bytes obtain no bonus on recall scores. Due to full-length F1 makes the evaluation result unbiased to summary length, we set no limitation to headline length when testing on Gigaword test set. 

For Chinese, we report full-length F1 scores ($R1_{F1}$, $R2_{F1}$ and $RL_{F1}$) following previous works~\cite{lcsts,copynet}. We set no length limitation on Chinese experiments either.

\subsection{Baseline Systems}
\label{sec:baseline}


\subsubsection{English Baseline systems}

\textbf{TOPIARY}~\cite{zajic2004bbn} is the winner system of DUC2004 Task-1. It utilizes linguistic-based sentence compression method and unsupervised topic detection at the same time.

\textbf{MOSES+} \cite{rush2015neural} generates headlines based on MOSES, a widely-used phrase-based machine translation system \cite{koehn2007moses}. It also enlarges the  phrase table and uses MERT to improve the quality of generated headlines.

\textbf{ABS} and \textbf{ABS+} \cite{rush2015neural} are both attention-based neural models that generate short summary for given sentence. The difference is that ABS+ extracts additional n-gram features at word level to revise the output of ABS model.

\textbf{RAS-Elman} and \textbf{RAS-LSTM}~\cite{rush2016} both utilize convolutional encoders that take input words and word position information into account. They also make use of attention-based decoders. The differernce is that, RAS-Elman selects Elman RNN~\cite{elman1990} as decoder, while RAS-LSTM selects long short term memory architecture ~\cite{hochreiter1997long}.

\textbf{BWL}, namely big-words-lvt2k-lsent ~\cite{CoRR} implements a trick that restricts the vocabulary size at the decoder end, by means of constructing the vocabulary of documents in each mini-batch respectively~\cite{jean2015}.

All the English baseline systems listed above except TOPIARY utilize Gigaword dataset for training, as described in Section \ref{sec:enData}.

\subsubsection{Chinese Baseline systems}

\textbf{RNN-context}~\cite{lcsts} is a simple character based encoder-decoder architecture that takes the concatenation of all hidden states at the encoder end as the input of decoder end.

\textbf{COPYNET} \cite{copynet} incorporates copying mechanism into sequence-to-sequence framework, which replicates certain segments from the input sentence into the output sentence.

\subsection{Implementation Details}

In MLE, the word embeddings are randomly initialized and then updated during training. In MRT, we initialize model parameters using the optimized parameters learned from NHG with MLE. For English models, we set the word embedding dimension to 620, the hidden unit size to 1,000 and the vocabulary size to 30,000. The corresponding values for Chinese models are 400, 500 and 3,500 respectively. In particular, the size of subset $\mathcal{S}(\mathbf{x};\theta)$ in Eq.(\ref{eq:loss}) has a great impact on the performance. When the size is too small, the sampling will not be sufficient. When the size is too large, the learning time will grow correspondingly. In this paper, we set the size to $100$ to achieve a trade-off between effectiveness and efficiency. These samples are drawn from the probability distribution of generated headlines by the up-to-date NHG model\footnote{An alternative subset building strategy is to choose top-$k$ headlines. Considering the efficiency and parallel architecture of GPUs, we opt sampling. }. We use AdaDelta~\cite{Zeiler2012ADADELTA} to adapt learning rates in stochastic gradient descent for both MLE and MRT. We utilize no dropout or regularization, but we take gradient clipping during training and the training is early stopped based on DUC2003 data. All our models are trained on GeForce GTX TITAN GPU. For NHG+MLE on the English dataset, it takes about $2.5$ hours for each $10,000$ iterations, For NHG+MRT, it takes about $3.75$ hours. During testing, we use beam search of size 10~\cite{rush2016} to generate headlines.
	
	

\subsection{Choices of Model Setup}

In the training process of NHG model, there are several significant factors that greatly influence the performance, such as the choice of distance measure in loss function and the treatment of unknown words. To determine the most appropriate choices of model setup, we investigate the effects of these factors on the development set respectively.

\subsubsection{Effect of Distance Measure}\label{sec:effect_en}

\begin{table}[!t]
	\centering\small
	\begin{tabular}{c|c|c|c|c}
		\hline
		\multicolumn{2}{c|}{}& \multicolumn{3}{c}{evaluation metric} \\
		\hline
		criterion 	& loss & R1  		& R2 		& RL \\
		\hline
		MLE 	& N/A 	& 23.70 		& 7.85 		& 21.20 \\
		\hline
		& $-$R1 	& {\bf 28.81} 	& 9.58  		&{\bf 25.31}  \\
		MRT 	& $-$R2 	& 26.94  		& 9.56 		&24.01\\
		& $-$RL 	& 28.19 		& {\bf 9.64 }	&25.02\\
		\hline
	\end{tabular}
	\setlength{\abovecaptionskip}{2pt}
	\caption{Effects of distance measures on the English validation set. $-$R1, $-$R2 and $-$RL represent the opposite value of ROUGE-1, ROUGE-2 and ROUGE-L respectively.}\label{tab:duc2003valid}
\end{table}
		
\begin{table}[!t]
	\centering\small
	\begin{tabular}{l|c|c|c|c}
		\hline
		\multicolumn{2}{c|}{}& \multicolumn{3}{c}{evaluation metric} \\
		\hline
		criterion & loss & R1  & R2 & RL \\
		\hline
		MLE 		& N/A 	& 24.61		& 8.52 		& 22.00 \\
		\hline
		& $-$R1 	& {\bf 29.84} 	& 10.24  		& {\bf 26.33}  \\
		MRT 	& $-$R2 	& 27.97  		& 10.18 		&  24.99\\
		& $-$RL 	& 29.18		& {\bf 10.44 }	& 25.88\\
		\hline
	\end{tabular}
	\setlength{\abovecaptionskip}{2pt}
	\caption{Effects of using different distance measures on the English test set. }\label{tab:duc2004valid}
\end{table}
			
As described in Section \ref{sec:rouge}, the distance $\Delta(\mathbf{y'},\mathbf{y})$ in the loss function is computed by the negative value of ROUGE. We investigate the effect of utilizing various distance measures in MRT. Table~\ref{tab:duc2003valid} shows the experiment results on English development set using different evaluation metrics. We find that all NHG+MRT models consistently outperform NHG+MLE, which indicates that the MRT technique is robust when loss function varies. $-$R1$_{R}$ statistically brings significant improvement for all evaluation metrics over MLE, one possible reason is that R1 score correlates well with human judgement~\cite{lin2004looking}. Hence we decide to utilize $-$R1$_{R}$ as the default semantic measure in the experiments (e.g., in Section \ref{sec:eva_result}). In addition,  Table ~\ref{tab:duc2004valid} shows that this argument is still valid on DUC2004.

\subsubsection{Effect of UNK Post Processing}

In the training procedure of NHG model, a common experiment setup is to keep a fixed size of vocabulary on both input and output side. These vocabularies are usually shortlists that only contain most frequent words, so that the out of vocabulary words are usually mapped to a special token ``UNK''. 

There are three typical post-processing methods to deal with UNK tokens. A simplest way is to ignore them, and we denote it as \textbf{Ignore}. Another way~\cite{jean2015} is to copy words from original input directly, and we denote it as \textbf{Copy}. The third way is to replace the unknown words according to a dictionary built upon the whole training set, and we denote it as \textbf{Mapping}. We conduct experiments on the English development set to investigate the performance of these methods. The fixed vocabulary size in NHG model is set to 30,000. Experiment results shown in table~\ref{tab: unk} indicate that the ``Copy'' method performs the best among three methods and generally improves the original model. Hence, we decide to utilize it as the default post processing method in our experiments on the test set.

\begin{table}[!t]
	\centering\small
	\begin{tabular}{l|c|c|c}
		\hline
		& $R1_{R}$ & $R2_{R}$ & $RL_{R}$\\
		\hline
		Original  & 28.08 & 9.19 & 25.00 \\\hline
		Ignore & 28.81 & 9.58 & 25.31 \\
		Copy  & \textbf{29.68} & \textbf{9.98} & \textbf{25.94} \\
		Mapping & 29.62 & 9.94 & 25.91 \\
		\hline
	\end{tabular}
	\setlength{\abovecaptionskip}{2pt}
	\caption{\small Effect of using different UNK Post Processing methods on English development set.}
	\label{tab: unk}
\end{table}

\begin{table}[!t]
	\centering\small
	\begin{tabular}{l|r|r|r}
		\hline
		& $R1_{R}$ & $R2_{R}$ & $RL_{R}$\\
		\hline
		Input-only &  27.17 & 8.98 & 23.96 \\
		\hline
		Extended-input & 28.08 & 9.19 & 24.50 \\
		\hline
		Full-vocab & \textbf{29.68} & \textbf{9.98} & \textbf{25.94} \\
		\hline
	\end{tabular}
	\setlength{\abovecaptionskip}{2pt}
	\caption{\small Effect of using different restrictions of output vocabulary on English development set.}
	\label{tab: trick}
\end{table}

\begin{table*}[t]
	\centering
	\small
	\begin{tabular}{l|c|l|c|c|c|c|c|c}
		\hline
		\hline
		\multirow{2}{*}{System}  & \multirow{2}{*}{Training} &  \multirow{2}{*}{Model Architecture}& \multicolumn{3}{c|}{DUC-2004} & \multicolumn{3}{c}{Gigaword} \\\cline{4-9}
		
		&  & & $R1_{R}$ & $R2_{R}$ & $RL_{R}$ & $R1_{F1}$ & $R2_{F1}$ & $RL_{F1}$ \\
		\hline
		\multicolumn{9}{c}{{\em Non-neural systems}} \\
		\hline
		TOPIARY	& \multirow{2}{*}{$-$}		& Linguistic-based	& 25.12 & 6.46 & 20.12 & - 	& - 	& - \\
		MOSES+	& 					& Phrase-based			& 26.50 & 8.13 & 22.85 & - & - & -\\
		\hline
		\multicolumn{9}{c}{{\em Neural systems}} \\
		\hline
		ABS 			& \multirow{5}{*}{MLE}	&Attention-based enc + NNLM& 26.55 & 7.06 & 22.05 & 29.55 & 11.32 & 26.42  \\
		ABS+ 		&					&ABS + Extractive tuning & 28.18 & 8.49 & 23.81 & 29.76 & 11.88 & 26.96\\
		RAS-Elman 	&					&CNN enc + Elman-RNN dec& 28.97 & 8.26 & 24.06 & 33.78 & 15.97 & 31.15\\
		RAS-LSTM	&					&CNN enc + LSTM dec& 27.41 & 7.69 & 23.06 & 32.55 & 14.70 & 30.03\\
		BWL &			&G-RNN enc + G-RNN dec + trick & 28.35 & 9.46 & 24.59 & 33.17 & 16.02 & 30.98\\
		\hline
		\multirow{2}{*}{{\em this work}}  & MLE &	G-RNN enc + G-RNN dec	&24.92 	& 8.60 & 22.25 & 32.67 & 15.23 & 30.56 \\
		& MRT &	G-RNN enc + G-RNN dec	&\textbf{30.41} &\textbf{10.87}& \textbf{26.79}&\textbf{36.54} & \textbf{16.59} & \textbf{33.44}\\
		\hline\hline
	\end{tabular}
	\setlength{\abovecaptionskip}{2pt}
	\caption{\small Comparison with baseline systems on DUC-2004 and Gigaword English test sets. G-RNN stands for Gated Recurrent Neural Networks, enc and dec are shorts for encoder and decoder respectively.} \label{duc2004}
\end{table*}
	
\begin{table*}[t]
	\small
	\centering
	\begin{tabular}{l|c|l|c|c|c}
		\hline
		\hline
		\multirow{2}{*}{System}  & \multirow{2}{*}{Training} & \multirow{2}{*}{Model} & \multicolumn{3}{c}{LCSTS}\\\cline{4-6}
		
		&  & & $R1_{F1}$ & $R2_{F1}$ & $RL_{F1}$ \\
		\hline
		RNN-context 		& \multirow{2}{*}{MLE}	&G-RNN enc + G-RNN dec + minimum length	& 29.9 & 17.4 & 27.2  \\
		COPYNET		&		&G-RNN enc + G-RNN dec + Copy mechanism& 35.0 & 22.3 & 32.0 \\
		\hline
		\multirow{2}{*}{{\em this work}}  & MLE &	G-RNN enc + G-RNN dec &	34.9  & 23.3 & 32.7  \\
		& MRT &	G-RNN enc + G-RNN dec &	\textbf{38.2}  & \textbf{25.2} & \textbf{35.4}\\
		\hline\hline
	\end{tabular}
	\setlength{\abovecaptionskip}{2pt}
	\caption{\small
		Comparison with baseline systems on Chinese test set. Note that the RNN-context has the same model architecture as ours. But they set a minimum length limit when decoding.}
	\label{tab:lcsts}
\end{table*}

\subsection{Evaluation Results}
\label{sec:eva_result}


Table \ref{duc2004} shows the evaluation results of headline generation on different English test sets. The baseline systems are introduced in Section \ref{sec:baseline}. These results indicate that NHG model with MLE achieves comparable performance to existing headline generation systems. Moreover, replacing MLE with MRT significantly and consistently improves the performance of NHG model, and outperforms the state-of-the-art systems on both test set.

Similar results can be observed from the experiment results on Chinese headline generation task as well, as shown in Table \ref{tab:lcsts}\footnote{The MRT result reported here is obtained by taking the negative F1 score of ROUGE-1 as loss fuction. Several realted experiment results are not given due to the length limit.}. NHG with MRT improves the ROUGE scores up to over $3$ points compared with baseline systems. We also notice that MLE model is already better than ~\cite{lcsts} and comparable to ~\cite{copynet}. This indicates that a character based model indeed performs good on Chinese summary task. Moreover, when evaluating with F1 scores, longer summaries would be penalized and get lower scores. 


\begin{figure}[h]
	\centering
	\includegraphics[width=7cm]{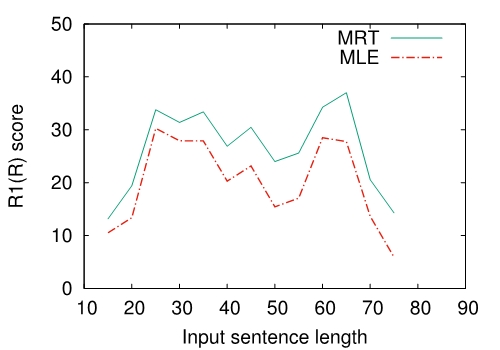}
	\setlength{\abovecaptionskip}{2pt}
	\caption{\small Recall scores of ROUGE-1 on DUC-2004 test set over various input lengths.}
	\label{fig:sent_len}
\end{figure}
Figure~\ref{fig:sent_len} shows the R1 scores of headlines generated by NHG+MLE and NHG+MRT on the English dataset with respect to input lengths.  As we can see, NHG+MRT consistently improves over NHG+MRT for all lengths.

\begin{table*}[t]
	\centering \small
	\begin{tabular}{l|p{13.5cm}}
		\hline \hline
		\textbf{Article (1)}: 	& Jose Saramago became the first writer in Portuguese to win the Nobel prize for literature on Thursday , his personal delight was seconded by a burst of public elation in his homeland .\\
		\textbf{Reference}: 	& Jose Saramago becomes first writer in Portuguese to win Nobel prize for literature \\
		\textbf{NHG+MLE}: 	& Portuguese becomes Portuguese president to win the nobel prize for literature \\
		\textbf{NHG+MRT}: 	& Jose Saramago is the first writer in the Portuguese language to win Nobel \\ 
		\hline
		\textbf{Article (2)}: 	& A slain Russian lawmaker was honored Tuesday as a martyr to democratic ideals in a stately funeral service in which anger mingled freely with tears .\\
		\textbf{Reference}: 	& Russian lawmaker buried beside greats; mourned as martyr; killers unknown.\\
		\textbf{NHG+MLE}: 	& Slain Russian lawmaker remembered as martyr to democracy ( Moscow ) \\
		\textbf{NHG+MRT}: 	& Slain Russian lawmaker honored as martyr in stately funeral service\\
		\hline
		\textbf{Article (3)}: 	& Voting mainly on party lines on a question that has become a touchstone in the debate over development and preservation of wilderness , the Senate on Thursday approved a gravel road through remote wildlife habitat in Alaska . \\
		\textbf{Reference}: 	& Senate approves 30-mile road in Alaskan wilderness; precedent? veto likely.\\
		\textbf{NHG+MLE}: 	& US senate passes law allowing road drilling in Alaska , Alaska \\
		\textbf{NHG+MRT}: 	& Senate passes gravel road through Alaska wildlife habitat in Alaska\\
		\hline\hline
	\end{tabular}
	\setlength{\abovecaptionskip}{2pt}
	\caption{\small
		Examples of original articles, reference headlines and generated outputs by different training strategy on DUC-2004 test set.}
	\label{duc2004example}
\end{table*}

To reduce the computation complexity when training NHG model, a possible approach is to restrict the size of vocabulary for generated headlines. There are three typical methods to deal with the size. The first one is to restrict the output words of headline within the input sentence, denoted as \textbf{Input-only}. The second one is to construct an extended vocabulary that includes similar words with those appear in the input sentence, denoted as \textbf{Extended-input}. The third one is to use full vocabulary, denoted as \textbf{Full-vocab}. Table \ref{tab: trick} illustrates the experiment results of using different restrictions on the English development set. The extended vocabulary is constructed by collecting 100 nearest neighbors for each input word, according to pre-trained Google-News word vectors~\cite{Mikolov2013Distributed}. We observe that the ``Input-only'' and "Extened-input" achieve comparable performance while using hundreds of times less vocabulary. It indicates that it is feasible to utilize these tricks to train NHG model much more efficiently.

\subsection{Case Study}

We present several examples for comparison as shown in Table \ref{duc2004example}. We can observe that: (1) NHG with MRT is generally capable of capturing the core information of an article. For example, the main subject in Article 1 is ``Jose Saramago''. NHG+MRT can successfully find the correct topic and generate a headline about it, but NHG+MLE failed. (2) When both systems capture the same topic, NHG+MRT can generate more informative headline. For Article 2, NHG+MLE generates ``remembered as'' when NHG+MRT generates `` honored as''. Considering the context, ``honored as'' would be more appropriate. (3) NHG+MLE usually suffer from generating duplicated words or phrases in headlines. As shown in Article 3, NHG+MLE repeats the phrase ``Alaska'' several times which leads to a semantically incomplete headline. NHG+MRT seems to be able to overcome this problem, benefitting from directly optimizing sentence-level ROUGE. 
				
\section{Related Work}
			
Headline generation is a well-defined task standardized in DUC-2003 and DUC-2004. Various approaches have been proposed for headline generation: rule-based, statistical-based and neural-based.
			
The rule-based models create a headline for a news article using handcrafted and linguistically motivated rules to guide the choice of a potential headline. Hedge Trimmer \cite{dorr2003hedge} is a representative example of this approach which creates a headline by removing constituents from the parse tree of the first sentence until it reaches a specific length limit. 
Statistical-based methods make use of large scale training data to learn correlations between words in headlines and articles~\cite{banko2000headline}. The best system on DUC-2004, TOPIARY \cite{zajic2004bbn} combines both linguistic and statistical information to generate headlines. There is also method make use of knowledge bases to generate better headlines. 
With the advances of deep neural networks, there are growing works that design neural networks for headline generation. \cite{rush2015neural} proposes an attention-based model to generate headlines. 
\cite{filippova2015sentence} proposes a recurrent neural network with long short term memory (LSTM)~\cite{hochreiter1997long} for headline generation. 
\cite{copynet} introduces copying mechanism into encoder-deconder architecture inspired by the Pointer Networks~\cite{pointer}. 

In this work, we propose the NHG model realized by a bidirectional recurrent neural network with gated recurrent units. We also propose to apply minimum risk training (MRT) to optimize parameters of NHG model. MRT has been widely used in machine translation \cite{och2003minimum,smith2006minimum,gao2014learning,shen2015minimum}, but less been explored in document summarization. To the best of our knowledge, this work is the first attempt to utilize MRT in neural headline generation.
			
\section{Conclusion and Future Work}
			
In this paper, we build an end-to-end neural headline generation model, which does not require heavy linguistic analysis and is fully data-driven. We apply minimum risk training for model optimization, which effectively incorporates sentence-wise information by taking various evaluation metrics into consideration. Evaluation result shows that NHG with MRT achieves significant and consistent improvements on both English and Chinese datasets, as compared to state-of-the-art baseline systems including NHG with MLE. There are still many open problems to be explored as future work: (1) Besides article-headline pairs, there are also rich plain text data not considered in NHG training. We will investigate the probability of integrating these plain texts to enhance NHG for semi-supervised learning. (2) We will investigate the hybrid approach of incorporating NHG with other successful headline generation approaches like sentence compression models. 
			
				
\bibliography{acl2016}
\bibliographystyle{acl2016}

\end{document}